\documentclass[journal]{IEEEtran}
\usepackage[T1]{fontenc}
\usepackage{graphicx}
\usepackage{times}
\usepackage{helvet}
\usepackage{courier}
\usepackage{amsmath}
\usepackage{algorithmic}
\usepackage{csquotes} 
\usepackage{color}
\usepackage{paralist}
\usepackage{amssymb}
\usepackage{indentfirst}
\usepackage{subfigure}
\usepackage{float}
\usepackage{multirow}
\usepackage{cite}
\usepackage{mathrsfs}

\usepackage{mathrsfs} 
\usepackage{amsfonts}
\usepackage{todonotes}
\usepackage{pgfplots} 
\usepackage{epstopdf}
\usepackage{epsfig}
\pgfplotsset{compat=newest}
\usepackage{color}
\definecolor{forestgreen}{RGB}{0,139,69}
\usepackage{multirow}
\usepackage{xcolor}
\definecolor{citecolor}{HTML}{0071bc}
\usepackage{tabularray}
\usepackage{algorithm}
\usepackage{amssymb}
\usepackage{textcomp} 
\usepackage{siunitx}  
\usepackage[colorlinks, linkcolor=red,  anchorcolor=blue, citecolor=citecolor]{hyperref} 

\usepackage{xcolor}
\definecolor{SeaGreen4}{RGB}{0,205,102} 
\definecolor{SlateBlue}{RGB}{106,90,205} 
\definecolor{DarkRed}{RGB}{178,34,34} 
	
\usepackage[switch]{lineno}

\usepackage{amsmath}        
\usepackage{array} 
\usepackage{booktabs}
\usepackage{hhline}

\usepackage{textcomp,booktabs}
\usepackage{amssymb}
\usepackage{pifont}

\usepackage{makecell}

\usepackage{colortbl}
\definecolor{mygray}{gray}{.9}
\definecolor{mypink}{rgb}{.99,.91,.95}
\definecolor{mycyan}{cmyk}{.3,0,0,0}

\begin{document}

\title{  Adversarial Attack for RGB-Event based Visual Object Tracking  }   

\author{Qiang Chen, Xiao Wang*, \emph{Member, IEEE}, Haowen Wang*, Bo Jiang, Lin Zhu, Dawei Zhang, Yonghong Tian, \emph{Fellow, IEEE}, Jin Tang
\thanks{$\bullet$ Qiang Chen, Xiao Wang, Haowen Wang, Bo Jiang, and Jin Tang are with the School of Computer Science and Technology, Anhui University. (email: e23301220@stu.ahu.edu.cn, \{xiaowang, wanghaowen, jiangbo, tangjin\}@ahu.edu.cn)}       
\thanks{$\bullet$ Lin Zhu is with Beijing Institute of Technology, Beijing, China (email: linzhu@pku.edu.cn)} 
\thanks{$\bullet$ Dawei Zhang is with Zhejiang Normal University (email: davidzhang@zjnu.edu.cn)} 
\thanks{$\bullet$ Yonghong Tian is with Peng Cheng Laboratory, Beijing, China. National Key Laboratory for Multimedia Information Processing, School of Computer Science, Peking University, China. School of Electronic and Computer Engineering, Shenzhen, Graduate School, Peking University, China. (email: yhtian@pku.edu.cn)} 
\thanks{* Corresponding Author: Xiao Wang, Haowen Wang}  
}

\markboth{ IEEE Transactions on ***, 2025 } 
{Shell \MakeLowercase{\textit{et al.}}: Bare Demo of IEEEtran.cls for IEEE Journals}

\maketitle

\begin{abstract}
Visual object tracking is a crucial research topic in the fields of computer vision and multi-modal fusion. Among various approaches, robust visual tracking that combines RGB frames with Event streams has attracted increasing attention from researchers. While striving for high accuracy and efficiency in tracking, it is also important to explore how to effectively conduct adversarial attacks and defenses on RGB-Event stream tracking algorithms, yet research in this area remains relatively scarce.
To bridge this gap, in this paper, we propose a cross-modal adversarial attack algorithm for RGB-Event visual tracking. 
Because of the diverse representations of Event streams, and given that Event voxels and frames are more commonly used, this paper will focus on these two representations for an in-depth study. Specifically, for the RGB-Event voxel, we first optimize the perturbation by adversarial loss to generate RGB frame adversarial examples. 
For discrete Event voxel representations, we propose a two-step attack strategy, more in detail, we first inject Event voxels into the target region as initialized adversarial examples, then, conduct a gradient-guided optimization by perturbing the spatial location of the Event voxels. For the RGB-Event frame based tracking, we optimize the cross-modal universal perturbation by integrating the gradient information from multimodal data. We evaluate the proposed approach against attacks on three widely used RGB-Event Tracking datasets, i.e., COESOT, FE108, and VisEvent. Extensive experiments show that our method significantly reduces the performance of the tracker across numerous datasets in both unimodal and multimodal scenarios. 
The source code will be released on \url{https://github.com/Event-AHU/Adversarial_Attack_Defense}
\end{abstract}


\IEEEpeerreviewmaketitle

\section{Introduction}

\IEEEPARstart{V}{isual} object tracking, as a fundamental task in the field of computer vision, has the core objective of achieving continuous spatio-temporal localization of target objects in an image sequence. Visual object tracking plays an important role in critical applied fields such as autonomous driving~\cite{chen2024end,leon2019review,luo2021exploring}, intelligent surveillance~\cite{dsouza2022artificial,yang2023cooperative}, and multimodal human-computer interaction~\cite{alzubi2025multimodal,islam2021cybersickness}. Conventional tracking architectures mainly focus on RGB imaging modalities that utilize dense photometric and texture information for object tracking~\cite{Zhang2021Multi-domain,chen2023seqtrack,hong2024onetracker}. However, these traditional methods often suffer significant performance degradation under challenging conditions, such as high-speed target object displacement, persistent occlusion, and drastic changes in environmental illumination. In recent years, breakthroughs in the field of neuromorphic sensing have introduced the Event camera as a bionic vision sensor~\cite{lichtsteiner2008128}. Event cameras encode pixel brightness changes as sparse tuples of spatio-temporal events $(x, y, t, p)$ in an asynchronous manner. The differential sensing paradigm of the Event camera achieves unprecedented {\textmu s}-level temporal resolution, 120 dB high dynamic range, and excellent motion blur immunity. 
Inspired by these features, some researchers have begun to explore multimodal fusion methods~\cite{tang2022revisiting,zhu2023cross,gehrig2020eklt} that combine the RGB image's rich appearance modeling with the event data's high temporal resolution motion dynamics, thereby advancing the innovation of visual tracking and enhancing the robustness of object tracking models in complex dynamic environments. 

\begin{figure} 
    \centering
    \includegraphics[width=\columnwidth]{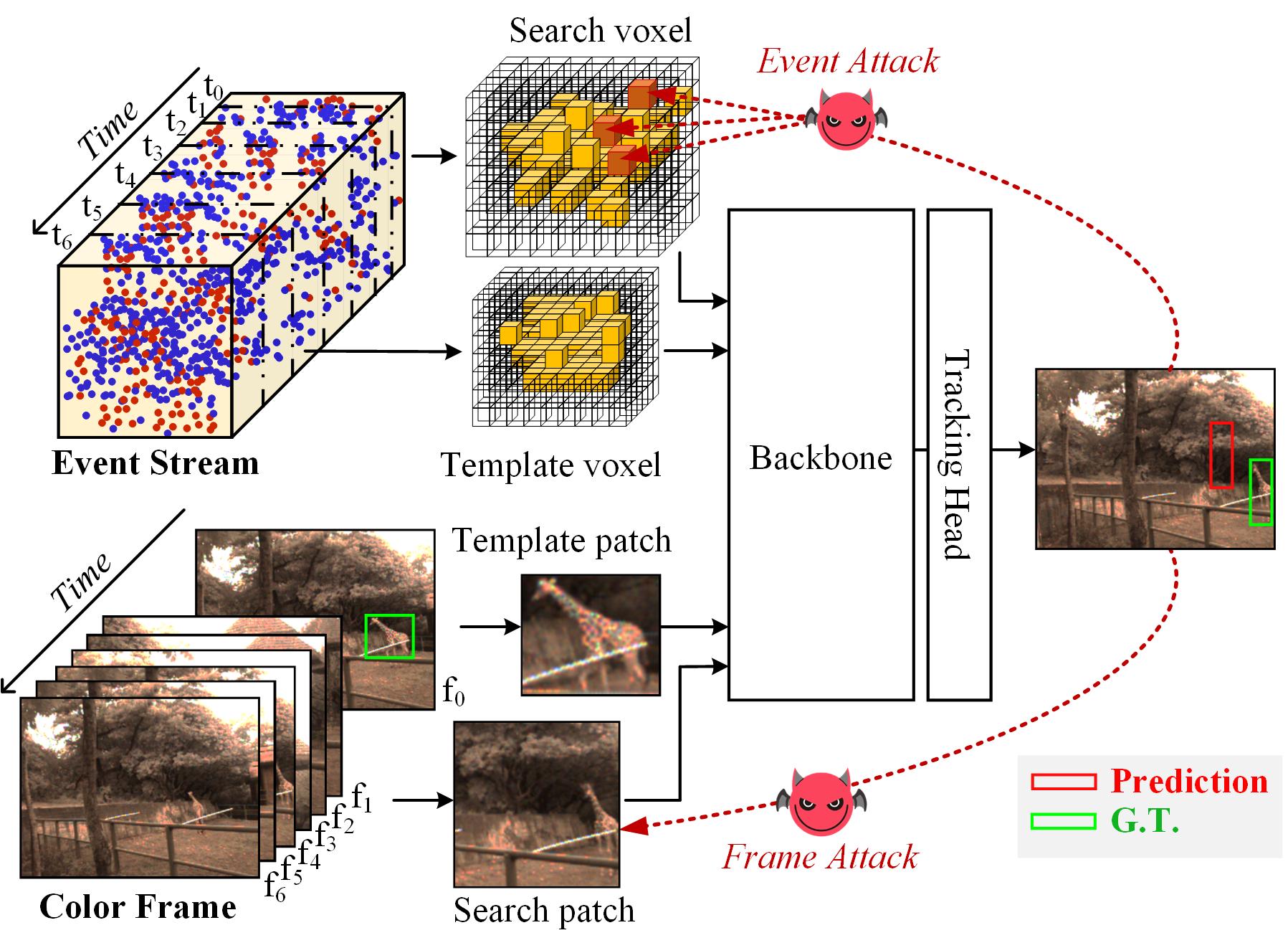} 
    \caption{An illustration of adversarial attacks for RGB-Event visual object tracking.} 
    \label{fig:rgbe_adv_frame_framework}
\end{figure}

Although multimodal visual object tracking(e.g., RGBE~\cite{sun2024reliable,shao2025tenet}, RGBT~\cite{wang2024temporal,wang2021adaptive,gao2025tvtracker,bunyak2007geodesic,xiao2016compressive,zhang2014fast}, and RGBD-based tracking~\cite{zhu2023rgbd1k,yan2021depthtrack,muller2021seeing,yan2021depthtrack,kart2019object,zhang2020occlusion}) has significantly advanced in recent years, there is still uncertainty about its security under adversarial instances, which brings potential risks for security-sensitive applications~\cite{dong2019efficient}. Adversarial attacks fool deep learning models by introducing nearly imperceptible perturbations in the input data. Currently, a lot of adversarial attack methods focus on image classification~\cite{szegedy2014intriguing,zeng2019adversarial,moosavi2017universal} and object detection~\cite{xie2017adversarial,gupta2021adversarial,lu2017adversarial} tasks, while there is relatively little research on adversarial attacks targeting object tracking in continuous image sequences. RGB-Event visual object tracking has three unique adversarial challenges compared to image classification~\cite{papernot2017practical,carlini2017towards,moosavi2016deepfool} and object detection~\cite{wei2019transferable,liu2018dpatch,xie2017adversarial} tasks:
\textit{1). Category-independence:} The labels of a visual object tracking task contain only bounding box information and do not involve the category information of the image. This prevents traditional attack methods for image classification and object detection from being directly and effectively applied to visual object tracking. Therefore, attack methods for visual object tracking need to fully utilize the parameter and structural information of the model to improve the attack success rate. 
\textit{2). Temporal Attack Consistency:} A successful perturbation must maintain adversarial consistency over consecutive frames to invert the temporal continuity inherent in the object trajectory. However, existing adversarial attack methods mainly focus on still images and lack research on attack methods for continuous image sequences.
\textit{3). Cross-modal perturbation complexity:} Attacking a multimodal tracking system requires coordinated adversarial noise interventions across heterogeneous data representations, and thus requires the design of collaborative perturbation strategies for different data modalities. 

Current adversarial algorithms for visual object tracking mainly focus on RGB frame-based network architectures, such as Jia et al.~\cite{jia2024robust}, who developed motion-aware perturbation propagation across video frames, and Chen et al.~\cite{guo2020spark}, who devised an attention-guided template destruction mechanism. 
However, there are still two critical research issues that need to be resolved. 
First, the inherent temporal continuity of event modalities and sparse event points fundamentally challenge perturbation optimization paradigms for RGB frame inputs. 
Second, although multimodal fusion~\cite{wang2024event,wu2024single,zhu2023cross} has been proven to be effective in improving model performance through cross-modal feature interactions, the potential risk of multimodal fusion frameworks inducing performance enhancement for adversarial attacks remains unexplored.  

To explore the adversarial robustness of Event streams and multimodal data in a visual object tracking framework, we propose the first adversarial attack algorithm for RGB-Event visual object tracking, as shown in Figure~\ref{fig:rgbe_adv_frame_framework}. For the input form of RGB-Event Voxel, we first optimize the perturbation by adversarial loss to generate adversarial examples of RGB frames. 
Subsequently, we combine regional Event voxel information injection with gradient-driven spatial optimization to generate adversarial examples for Event Voxel. 
For the RGB-Event frame, we optimize the generalized perturbation by integrating the gradient information from multimodal data across modalities. 
In addition, we propose an adversarial loss function based on the tracking loss function to generate more effective adversarial examples. 
Meanwhile, the temporal perturbation strategy dynamically propagates the historical perturbations to the subsequent attacks to disrupt the temporal consistency. 
Evaluations on benchmark datasets (e.g., COESOT~\cite{tang2022revisiting}) show that our approach significantly disrupts the accuracy of the tracking model, which for the first time reveals the vulnerability of Event data and multimodal visual tracking systems in adversarial scenarios.

To sum up, the key contributions of this work can be summarized as follows:

1). We propose a novel approach, which models the spatial coordinates of Event voxels to generate more effective adversarial examples, thereby exploring the robustness of Event streams under adversarial attacks for the first time. 

2). We first explore the adversarial robustness of RGB-Event multimodal tracking by adopting the PGD algorithm~\cite{madry2017towards}, which leverages adversarial loss to propagate cross-modal attacks.

3). We conduct a robustness analysis of Event streams under both continuous Event frames and discrete Event voxel representations, uncovering potential vulnerabilities in multimodal visual tracking systems.

\section{Related Works} 

\subsection{RGB-Event based Tracking}
RGB-Event single-object tracking aims to overcome the limitations of traditional methods in complex scenarios by leveraging the complementary advantages of RGB and Event modalities. 
However, this approach introduces new challenges in feature fusion, parameter efficiency, and spatiotemporal modeling. 
To address the issue of cross-modal feature integration, Tang et al.~\cite{tang2022revisiting} pioneered the use of an adaptive Transformer architecture to achieve aggregation of cross-modal information; Wang et al.~\cite{wang2023visevent} designed a cross-modal Transformer module to construct a spatiotemporal alignment mechanism; Zhang et al.~\cite{zhang2021object} converted Event data into temporal slices, solving the inefficiency in fusion through adaptive modality contribution; Wu et al.~\cite{wu2024single} established a shared latent space via low-rank decomposition, promoting a new paradigm for RGB-Event modality fusion. 
In addressing the problem of parameter redundancy, Hong et al.~\cite{hong2024onetracker} developed a pre-training and prompt fine-tuning paradigm to achieve lightweight cross-modal transfer; Zhu et al.~\cite{zhu2023visual} utilized dynamic prompt vectors to bridge heterogeneous feature spaces, significantly enhancing generalization capabilities under small sample conditions. 
For the modeling challenges posed by high-speed motion, Huang et al.~\cite{Huang2018Event} proposed a dual-guidance mechanism based on the spatiotemporal continuity of Event streams; Zhang et al.~\cite{Zhang2025Revisiting} introduced a cascaded network that integrates motion trajectories with appearance features, thereby overcoming the limitations of traditional appearance-based matching.

\subsection{Adversarial Attack} 
In the field of visual object tracking, adversarial attack research is categorized into white-box~\cite{chen2021unified, chen2020one, guo2021learning, jia2020robust, liang2020efficient, wiyatno2019physical, zhao2023pluggable, yan2020hijacking} and black-box attacks~\cite{jia2021iou, yin2024dimba} based on the attacker's prior knowledge, with core challenges focusing on developing attack strategies under the temporal dependencies and localization decisions of tracking tasks. White-box attacks achieve perturbation injection by leveraging model gradient information and spatiotemporal correlations. Representative approaches include Jia et al.'s ~\cite{jia2024robust} lightweight perturbation propagation mechanism that perceives cross-frame motion signals, achieving cumulative perturbation effects through spatiotemporal correlation modeling; Li et al.'s~\cite{li2021simple}  collaborative perturbation strategy across template and search regions to induce false localization decisions; Guo et al.~\cite{jia2021iou} proposed a spatially-aware online incremental attack framework, which achieves low-iteration convergence through spatiotemporal correlation modeling of historical perturbations to generate imperceptible adversarial examples for real-time object tracking. and Chen et al.'s~\cite{guo2020spark} dual-attention-driven confidence-feature joint optimization framework, which integrates spatiotemporal perturbation propagation with multi-objective loss functions to overcome dynamic scene adaptability bottlenecks. Black-box attacks aim at model-agnostic conditions, such as Yan et al.~\cite{chen2020one} proposed a Cooling-Shrinking Attack, which dynamically generates perturbations by integrating a temporal IoU decay mechanism with historical frame confidence, simultaneously achieving heatmap suppression and bounding box shrinkage. Huang et al.'s~\cite{huang2024context} context-guided attack strategy was constructed via historical state inference, and Zhou et al.'s~\cite{zhou2023only} design of a confidence-localization joint loss function based on a UNet generation architecture, enabling the creation of adversarial samples without internal model information by modeling tracker behavior patterns and output response characteristics. 

However, current research predominantly focuses on adversarial attacks within the RGB modality~\cite{liang2020efficient, wiyatno2019physical, sheng2024towards, nakka2020temporally, ding2021towards}, leaving the robustness of Event-stream modalities and RGB-Event multi-modal fusion frameworks largely unexplored. There is an urgent need to investigate cross-modal attack strategies and corresponding defense mechanisms to address future security challenges.

\section{Our Proposed Approach} 

\begin{figure*}[h]
    \centering
    \includegraphics[width=\textwidth]{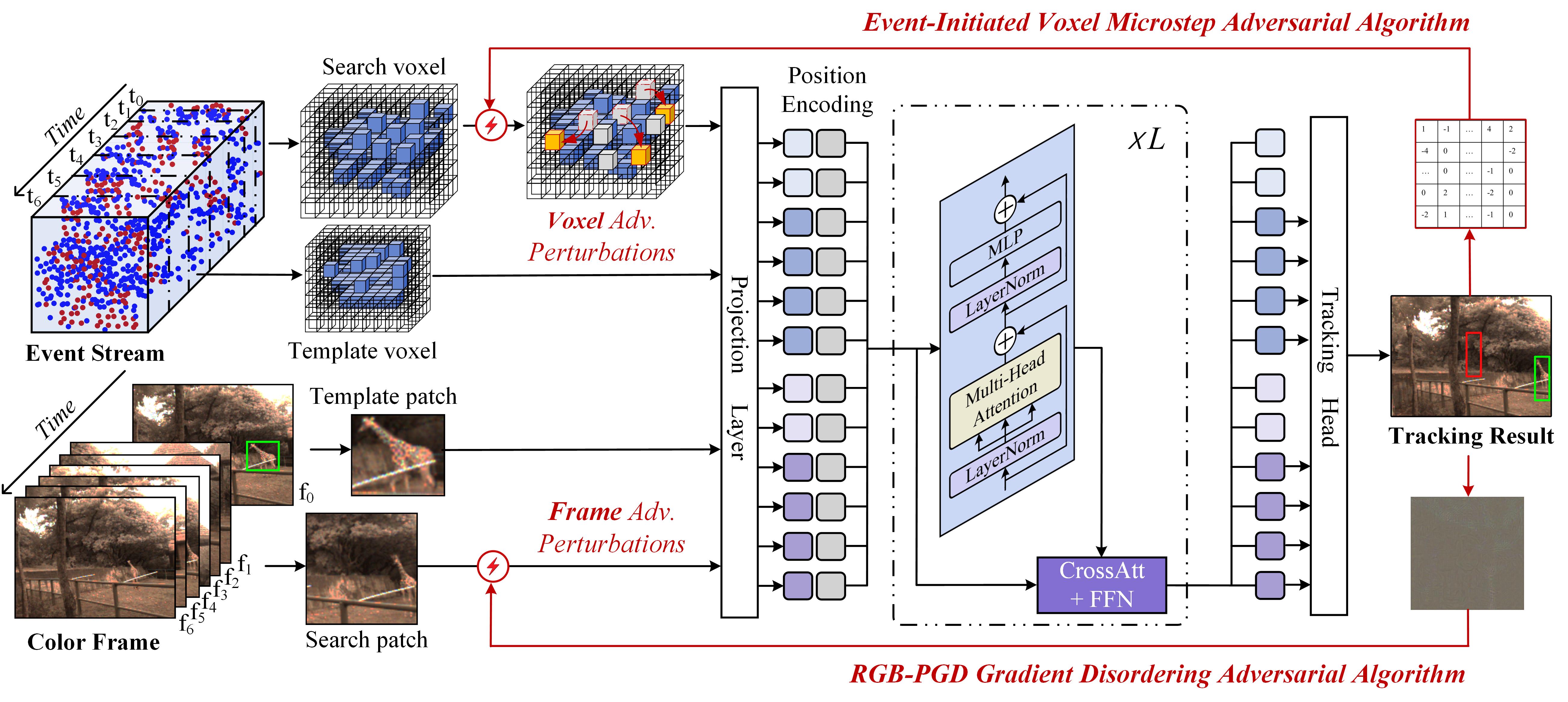}
    \caption{An overview of our proposed multi-modal adversarial attack framework for the RGB-Voxel based tracking.}
    \label{fig:rgbe_adv_voxel_framework}
\end{figure*}

\subsection{Overview}  
\label{sec:Overview}
In this study, we first explore the problem of adversarial robustness in RGB-Event visual object tracking and propose a multi-modal adversarial attack algorithm in response to the vulnerability of the multi-modal tracking model under adversarial perturbations. We individually design a target attack strategy for two typical representations of Event data, Event voxels and Event frames. For the RGB-Event voxel modality input, we adopt a two-step attack method: firstly, we generate the RGB frame adversarial samples based on the adversarial loss function; For the Event voxel, we initialize the adversarial samples in the target region through region injection to simulate the invalid voxel's spatial distribution, and optimize the spatial coordinates of Event voxels by using the PGD algorithm to generate the ultimate Event voxel adversarial sample. For the RGB-Event frame modality input, we adopt a cross-modal generalized perturbation strategy, which generates cross-modal generalized perturbation and RGB-Event frame adversarial samples by fusing the gradient information of RGB frames and Event frames. In addition, in order to keep the temporal consistency, we introduce a dynamic perturbation propagation mechanism, which passes the perturbation from history frames to the subsequent frames and systematically destroys the temporal correlation capability of the tracker. Finally, we construct an adversarial loss function based on the object tracking loss function to further generate more efficient adversarial samples.

\subsection{Problem Formulation} 
\label{sec:pro_define}
RGB-Event based visual tracking framework enables robust target localization in dynamic environments by leveraging spatio-temporal complementary cues through multi-modal perception.
At its core, visual tracking aims to establish dynamic spatio-temporal correlations between a target template and a search region.
The tracker localizes the target by performing discriminative template matching, effectively decoupling the target’s appearance variations from background interference within the search region.

Given an Event stream $\mathcal{E}$ and a RGB frame $\mathcal{I}$, the Event data is either discretized into a spatio-temporal voxel grid $V \in \mathbb{R}^{H \times W \times T}$ or accumulated into an Event frame $F \in \mathbb{R}^{H \times W}$. Based on these inputs, template patches $Z_r$ (RGB) and $Z_e$ (Event), as well as search patches $X_r$ and $X_e$, are extracted to construct a quaternion input ${Z_r, X_r, Z_e, X_e}$. This quaternion input undergoes positional encoding and cross-modal feature interaction before the fused search region features are fed into the tracking head for accurate target localization.

Adversarial attacks aim to cause model mispredictions by injecting imperceptible perturbations into input data. While adversarial attacks have been extensively explored in image classification and object detection, applying them to visual tracking presents distinct challenges. Unlike classification or detection, visual tracking involves template-based initialization, sequential frame prediction, and temporal propagation of bounding boxes, all of which complicate the design of effective attacks. 

\subsection{Event-to-Voxel Representation} \label{sec:Event voxelization}
Event data provides a sparse spatiotemporal representation of visual information through an asynchronous trigger mechanism. The core characteristic of an Event camera is its sensitivity to local brightness changes in the scene. Specifically, each pixel independently monitors the instantaneous difference in ambient brightness \( L(x,y,t) \). 
When \( \left| L(x,y,t) - L(x,y,t-\Delta t) \right| \) is over the contrast threshold \( C \), Event stream \( \mathcal{E} = \left\{ (x_{k}, y_{k}, t_{k}, p_{k}) \right\}_{k=1}^{N} \) is triggered. 
Event data exist as discrete spatiotemporal point sets with coordinates \( (x_k, y_k, t_k) \), which correspond to the spatiotemporal location of the brightness abrupt change, and polarity \( p_k \) encodes the physical characteristics of light intensity variation. 
Regions that are static or do not meet the threshold produce no data.

To preserve the temporal and spatial distribution characteristics of the Event stream, a temporal discretization window is applied to segment continuous events into discrete Event voxels.  
Specifically, fragments of the Event stream are extracted between two consecutive RGB frames. A sparse 3D voxel grid is then constructed by quantizing each event’s coordinates \( (x, y, t) \) based on a predefined voxel size, while also limiting the number of events assigned to each voxel. Each voxel is represented as \( V = \left[ V_x, V_y, V_z, V_f \right] \), where \( V_x, V_y, V_z \) indicate the voxel’s spatial-temporal location, and \( V_f \) denotes the aggregated feature computed by summing the polarities of all events within the voxel.

\subsection{Event Voxel-based Adversarial Perturbation} \label{sec:Unimodal Adversarial Perturbation on Event-Voxel VOT}

In order to generate adversarial examples of Event voxel in discrete space, we propose an initial perturbation injection strategy based on invalid voxel distribution statistics and a gradient-guided optimization method for spatial coordinates.

$\bullet$ \textbf{Initial Perturbation Injection}: Event voxels are used as input data to the object tracking model to predict the location of template patches within a search patch. 
In continuous event sequence object tracking, not every sequence has sufficient valid event information, and for those with insufficient event data, the tracking model uses zero padding.  
Accordingly, the initial perturbation injection strategy statistically quantifies the number of invalid voxels $N_v$ in the current search patch and subsequently injects $N_v$ voxels into a predefined target attack region as the initial adversarial examples.

$\bullet$ \textbf{Gradient-Guided Optimization}: Event voxel-based object tracking relies on the location information of events for localization. Therefore, we introduce perturbations to the $(x,y,t)$ coordinates of the initialized adversarial Event voxel examples and employ the PGD algorithm~\cite{madry2017towards} to generate adversarial examples. First, the gradient information is obtained by deriving the input Event voxel data through the adversarial loss function: 
\begin{equation}
    R_{xyt} = \frac{\partial L_{\text{adv}}}{\partial V^{'}},
\end{equation}
where $L_{\text{adv}}$ is the adversarial loss and $V^{'}$ is the input initialized Event voxel adversarial example. 
At each iteration, we update the perturbation in the direction of the loss function gradient. The iterative equation is as follows: 
\begin{equation}
    V_{m+1} = \text{Proj}_{{\epsilon}} \left( V_m + \alpha \cdot \text{sign}(R_{xyt}) \right)
\end{equation}
where $R_{\text{xyt}}$ represents the gradient information in the $(x,y,t)$ directions, $\alpha$ is the step size, and $m$ is the iteration, and $\epsilon$ is the maximum perturbation range. 
The projection operation adjusts the direction of the gradient to follow the boundary of the feasible region to satisfy the constraints.

$\bullet$ \textbf{Adversarial Loss Function}: 
The following loss function is typically used during tracker training:
\begin{equation}
\label{track loss}
\mathcal{L} = \lambda_1 \mathcal{L}_{\text{focal}}(y, y') + \lambda_2 \mathcal{L}_{L1}(b, \hat{b}) + \lambda_3 \mathcal{L}_{\text{GIoU}}(b, \hat{b})
\end{equation}
where $\mathcal{L}_{\text{focal}}$ mitigates foreground-background imbalance via adaptive weighting, $\mathcal{L}_{L1}$ penalizes center coordinate deviations, and $\mathcal{L}_{\text{GIoU}}$ enforces spatial consistency of bounding boxes, with $\lambda_i$ balancing their contributions.

In order to enhance the attack effect in discrete space, we propose the following adversarial loss function based on \label{track loss}:
\begin{equation}
\mathcal{L}_{adv}(V) = \sum_{z \in \mathcal{Z}} \sum_{t \in \mathcal{T}} \alpha_{z,t} \cdot e(z, y_t) - e(y, y_{\text{ori}})
\label{Single mode attack loss function}
\end{equation}
where $Z \in \{y, y_{\text{ori}}\}$, and $\mathcal{T} \in \{\text{target}, \text{true}\}$. $y$ denotes the adversarial prediction, $y_{\text{true}}$ the ground-truth label, $y_{\text{ori}}$ represents the original model's output on unperturbed inputs, and $y_{\text{target}}$ the target label, $e$ is the model that trained the tracker, $\alpha_{z,t}$ is a hyperparameter if t = $\text{target}$ is 1 otherwise -1.

The $\mathcal{L}_{\text{adv}}$ loss function minimizes the geometric distance between the adversarial sample and the predefined target position while maximizing the distance from the adversarial sample to both the ground-truth position and the initial prediction. The original model predictions are closer to the preset target position and further away from the ground-truth label position. Based on the above constraints, the final adversarial loss is formulated as a linear combination of $\mathcal{L}_{\text{focal}}$, $\mathcal{L}_{L1}$ and $\mathcal{L}_{\text{GIoU}}$.

$\bullet$ \textbf{Temporal Perturbation}: During the inference process of the object tracking model, the model achieves continuous prediction of the object bounding box by modeling the temporal information of the sequence. In order to effectively disrupt the temporal continuity, we adopt a perturbation optimization strategy based on temporal correlation:
\begin{equation}
V^{t}_{\text{adv}} = V^{t}_{\text{ori}} + (V^{t-1}_{\text{ori}} - V^{t-1}_{\text{adv}})
\label{eq:adv_update}
\end{equation}
where $V^{t-1}_{adv}$ and $V^{t}_{adv}$ denote adversarial examples computed at times t-1 and t; $V^{t-1}_{ori}$ and $V^{t}_{ori}$ denote original inputs recorded at times t-1 and t. We transfer the adversarial perturbation $\epsilon_{t-1}$ at time t-1 as the initial perturbation to time t. Before optimizing the adversarial example at moment t, $\epsilon_{t-1}$ is used as the starting point for optimization, which maintains the temporal consistency of the perturbation and significantly improves the effectiveness of the adversarial attack.

\subsection{Cross-Modal Attack Fusion for RGB-Event Voxel VOT} \label{sec:Cross-Modal Attack Fusion for RGB-Event Voxel VOT}

\begin{algorithm}[t]
\caption{Algorithm for cross-modal attack fusion against RGB-Event voxel VOT}
\label{rgb_event_voxel_attack}
\raggedright  
\textbf{Input}: Original RGB Frames ${I}$, Original Event Voxel $V=\{(x_k,y_k,t_k,p_k)\}_{k=1}^{N}$, Target location $S(x,y,w,h)$, Perturbation $\eta_{\text{rgb}}$, $\eta_{\text{event}}$, true label $gt$ \\
\textbf{Parameter}: Number of total iteration $M$, step size $\alpha$, perturbation size $\epsilon$,model parameters $\theta$  \\
\textbf{Output}: Adversarial RGB frames ${I}^{\ast}$, Adversarial Event voxel $V^{\ast}$
\begin{algorithmic}[1]
    \FOR{$t = 1$ to $T$}
        \STATE \textcolor{blue}{\% Initialize}
        \STATE Retrieve the RGB frame ${I}^{t}$ and Event voxel $V^{t}$ within the search space at the current temporal instance
        \STATE Get invalid number of voxel $N_v$ 
        \STATE ${V^{t}}' \leftarrow $ Add $N_v$ voxel in the target region $S(x,y,w,h)$
        \STATE Update ${I}^{t},{V^{t}}'$ according to $\eta_{\text{rgb}},\eta_{\text{event}}$ 
        \STATE \textcolor{blue}{\% Attack RGB frame} 
        \FOR{$i = 1$ to $M$}
            \STATE Compute $R_{rgb}=\nabla J({I_{M}^{t}},\theta)$ according to Equ.~\eqref{Single mode attack loss function}
            \STATE ${I}^{\ast} \leftarrow \text{Proj}_{\epsilon}\left({I_{M}^{t}} + \alpha \cdot \text{sign}(R_{rgb})\right)$  
        \ENDFOR
        \STATE Update $\eta_{\text{rgb}} \leftarrow {I^{\ast }} - I$
        \STATE \textcolor{blue}{\% Attack Event voxel} 
        \FOR{$i = 1$ to $M$}
            \STATE Compute $R_{event} = \nabla J({V^{t}_{M}}', \theta)$ according to Equ.~\eqref{Single mode attack loss function}
            \STATE $V^{\ast} \leftarrow \text{Proj}_{\epsilon}\left({V^{t}_{M}}' + \alpha \cdot \text{sign}(R_{event})\right)$
        \ENDFOR
        \STATE Update $\eta_{\text{event}} \leftarrow V^{\ast} - V$
    \ENDFOR
    \STATE Return $I^{\ast}$, $V^{\ast}$ 
\end{algorithmic}
\end{algorithm}

In the context of multi-modal inputs (RGB images and Event voxels), the performance of the object tracking model is significantly boosted. To further explore whether multi-modal inputs can strengthen the effectiveness of adversarial attacks, we propose a method to generate adversarial examples in a multi-modal shared space. Figure~\ref{fig:rgbe_adv_voxel_framework} illustrates our adversarial attack framework for the RGB-Event voxel visual object tracking model. First, we generate the adversarial examples for the RGB Frame based on the adversarial loss function~\eqref{Single mode attack loss function} using the following iterative equations:
\begin{equation}
I_{m+1} = \text{Proj}_\epsilon \left( I_m + \alpha \cdot \text{sign}(\frac{\partial L_{\text{adv}}}{\partial I_m}) \right)
\label{rgb_adv}
\end{equation}
where $\alpha$ denotes the step size, $\epsilon$ the perturbation budget, and $m$ the iterations.

Later, for the adversarial examples generation of Event voxels, we adopt the method described in Section~\ref{sec:Event voxelization}. Subsequently, the generated adversarial examples of RGB frames and Event voxels are input into the trained tracking model to obtain the final attack results. Algorithm~\ref{rgb_event_voxel_attack} describes in detail the overall process of generating the adversarial examples.
\begin{figure}[t]
    \centering
    \includegraphics[width=\columnwidth]{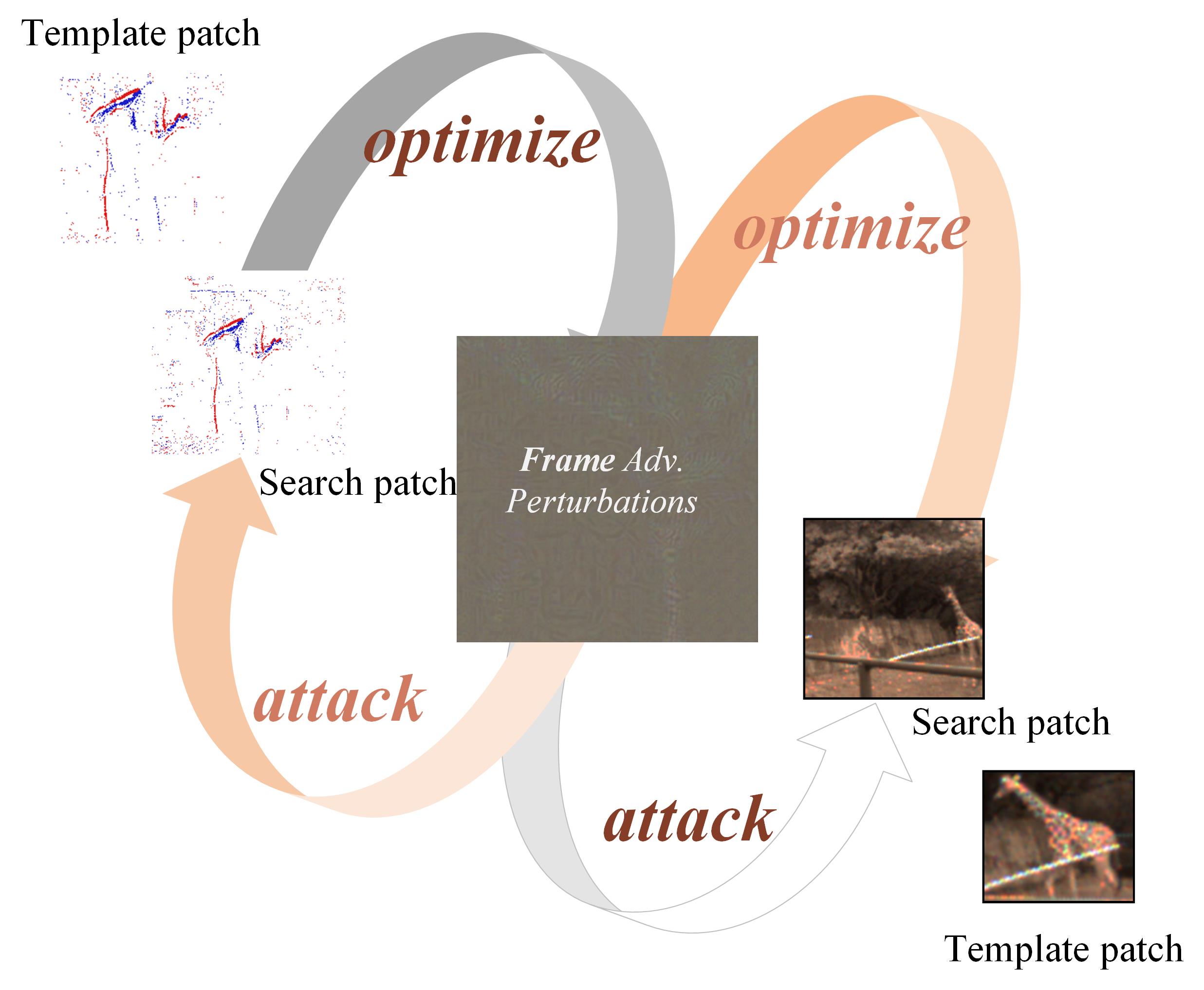} 
    \caption{An overview of our proposed multi-modal adversarial attack framework for the RGB-Event Frame based tracking.}
    \label{fig:rgbe_adv_frame_framework}
\end{figure}

\subsection{Multimodal Adversarial Attack on RGB-Event Frame VOT} \label{sec:Multimodal Adversarial Attack on RGB-Event Frame VOT}
Event streams are converted to discrete voxel representations by the methods in Section~\ref{sec:Event voxelization}. Event streams can be converted into image frame form by stacking on the coordinate axes. Current attack strategies mainly focus on the features of RGB frame modalities, overlooking the features of other modalities and the heterogeneity of inter-modal data distribution. Therefore, we propose a cross-modal collaborative attack method, which optimizes the perturbation by integrating the gradient information of multi-modal data to effectively disrupt the performance of the tracker. As shown in Figure~\ref{fig:rgbe_adv_frame_framework}, we adopt a two-step training process for the cross-modal universal perturbation. We first learn the universal perturbation $\eta$ based on the RGB frame modality and optimize $\eta$ by the adversarial loss function~\eqref{Single mode attack loss function}, the corresponding formula for this step is as follows:
\begin{equation}
\eta = \text{Clip} \left( \left | \alpha \cdot \text{sign}(\frac{\partial L_{\text{adv}}}{\partial I})\right |>\epsilon  \right)
\label{rgb_adv}
\end{equation}
where $\varepsilon$ is the perturbation boundary, we superimpose the RGB frame optimized $\eta$ on the Event frame to optimize $\eta$ in the second step.

In the second step, we further optimize the perturbation $\eta$ using the Event frame modality and stack the optimized $\eta$ to the RGB frame for the next perturbation optimization. Compared with the first step, the second step has differences in the initialization value and gradient update of the perturbation. The specific algorithmic process is shown in the Table ~\ref{alg:rgb_event_frame_attack}.

\begin{algorithm}[t]
\caption{Algorithms for adversarial attacks on RGB Event Frame-based visual object tracking}
\label{alg:rgb_event_frame_attack}
\raggedright
\textbf{Input}: Original RGB frames $I$, original Event frames $V$, pre-trained tracker $T$, target location $S(x,y,w,h)$, perturbation $\eta$, true label $gt$ \\
\textbf{Parameter}: number of total iteration $M$, step size $\alpha$, perturbation size $\epsilon$, and the model's parameters $\theta$ \\
\textbf{Output}: Adversarial RGB frames ${I^{*}}$ and Adversarial event frames ${V^{*}}$
\begin{algorithmic}[1]
    \FOR{$t = 1$ to $T$}
        \STATE \textcolor{blue}{\% Initialize}
        \STATE Get the RGB frames $I$ and Event frames $V$ in current search patch
        \STATE Initialize adversarial RGB frame ${I^{t}}' \leftarrow I^{t} + \eta$, Event frame ${V^{t}}' \leftarrow V^{t} + \eta$
        \FOR{$i = 1$ to $M$}
            \STATE \textcolor{blue}{\% Attack RGB frame}
            \STATE Compute $R_{rgb} = \nabla J({I^{t}_M}', \theta)$ according to Equ.~\eqref{Single mode attack loss function}
            \STATE ${I^{*}} \leftarrow \text{Proj}_{\epsilon}\left({I^{t}_M}' + \alpha \cdot \text{sign}(R_{rgb})\right)$
            \STATE Update $\eta \leftarrow {I^{*}} - I$
            \STATE \textcolor{blue}{\% Attack Event frame}
            \STATE Compute $R_{event} = \nabla J({V^{t}_M}', \theta)$ according to Equ.~\eqref{Single mode attack loss function}
            \STATE ${V^{*}} \leftarrow \text{Proj}_{\epsilon}\left({V^{t}_M}' + \alpha \cdot \text{sign}(R_{event})\right)$
            \STATE Update $\eta \leftarrow {V^{*}} - V$
        \ENDFOR
    \ENDFOR
    \STATE Return ${I^{*}}$,${V^{*}}$ and $\eta$
\end{algorithmic}
\end{algorithm}

\section{Experiments}  

\subsection{Dataset and Evaluation Metric}
We verify the effectiveness of our attack method on the COESOT~\cite{tang2022revisiting}, FE108~\cite{zhang2021object}, and VisEvent~\cite{wang2023visevent} datasets. A brief introduction to these datasets is given below. 

\textbf{1). COESOT Dataset.~}
The COESOT dataset provides strictly synchronized raw Event streams and RGB frames. The COESOT contains more than 30 categories of objects and is annotated with more than 10 complex attributes. In addition, COESOT provides multi-scale image capture with zoom lens technology and full source files.

\textbf{2). FE108 Dataset.~}
The FE108 dataset is an indoor tracking dataset captured by a grayscale Event camera DVS346 and provides event data and grayscale images in the scene. FE108 contains 108 videos in 21 object target categories covering fast motion scenes in low light conditions, in high dynamic range environments, and with and without motion blur on visual frames.

\textbf{3). VisEvent Dataset.~}
The VisEvent dataset contains 820 video sequences. 500 for training and 320 for testing. VisEvent covers challenging scenarios such as low-light conditions, high-speed motion, and motion blur. However, some sequences of VisEvent lack Event source files (*.aedat4), which limits their further development in the future. In our study, we use sequences containing source files converted into Event voxels for experiments.

To evaluate the performance of adversarial attack methods, we adopt the following metrics to measure the performance degradation of the tracker: 
\textit{1)~Precision Rate (PR).~} Percentage of frames with target center error below a threshold; 
\textit{2)~Normalized Precision Rate (NPR).~} Percentage of frames with normalized target center distance below a threshold; 
\textit{3)~Success Rate (SR).~} Area under the curve of the frames with IoU higher than a threshold.
Lower values of these metrics indicate a larger degradation in tracking performance, thus proving the effectiveness of the attack method.

\subsection{Implementation Details} 
In unimodal (RGB/Event) and multi-modal (RGB-Event frame/RGB-Event voxel) attack scenarios, we set the maximum perturbation margins to 10 and 8 pixels/offset unit and the perturbation steps to 1 and 1 pixel/offset unit, respectively. The number of iterations for the generation of the adversarial examples is 10. The parameters of the visual object tracking model remain frozen during the attack. We choose CEUTrack as the attacked visual object tracking model. CEUTrack is the first unified RGB-Event multi-modal tracking framework that fuses RGB frames and Event stream features by using a visual Transformer as the backbone network. Our experiments are conducted on a server with NVIDIA RTX-3090Ti GPUs based on the PyTorch~\cite{paszke2019pytorch} framework. More details can be found in our source code, which will be made available upon acceptance.

\subsection{Benchmarked Baselines} 
To assess the effectiveness of adversarial attacks on visual object tracking models using Event voxel, we conduct a comprehensive comparison with several state-of-the-art methods, including AE-ADV~\cite{lee2022adversarial} and DARE-SNN~\cite{yao2024exploring}  for Event streaming attacks. These methods were originally designed for image classification, and we transfer their algorithmic thoughts to the visual object tracking domain. To ensure the consistency of the multi-modal attacks, we adopt the PGD algorithm~\cite{madry2017towards} for all the attacks using image data during the multi-modal attacks. In addition, we also adopt FGSM (one-step perturbation attack)~\cite{goodfellow2014explaining} to adjust the attacks on Event and RGB data. Specific algorithmic details can be viewed in the Supplementary Material. Track loss function is an adversarial loss function~\eqref{Single mode attack loss function} that does not involve the loss term of the original tracking result, which is commonly used in adversarial attacks. We use it as a comparative method to optimize the loss function for adversarial samples. It is worth noting that the GradOpt algorithm is highly coupled with the Track loss function or Adv. loss function mentioned in Eq.~\eqref{Single mode attack loss function}.

\subsection{Comparison on Public Benchmark Datasets}  
\noindent $\bullet$ \textbf{Results on COESOT Dataset.~} As shown in Tables~\ref{sigle_modlity_attack_result} and~\ref{multimodal_attack_results}, our attack method shows significant attack performance in both unimodal and multi-modal attack scenarios. In the unimodal attack, the FGSM~\cite{goodfellow2014explaining} leads to a 4.6\%, 4.4\%, and 66.7\% decrease in the PR of Event voxels, Event frames, and RGB frames, respectively; AE-ADV~\cite{lee2022adversarial} and DARE-SNN~\cite{yao2024exploring} lead to a 3.4\% and 0.9\% decrease in the PR of Event voxels, respectively. In contrast, our method caused the PR of Event voxels, Event frames and RGB frames to decrease by 5.7\%, 54.2\% and 69\%, respectively.
In multimodal scenarios, the FGSM~\cite{goodfellow2014explaining} leads to a 34\% and 68.9\% decrease in the PR of RGB-Event voxel and RGB-Event frame, respectively, and AE-ADV~\cite{lee2022adversarial} and DARE-SNN~\cite{yao2024exploring} lead to a 69.2\% and 69.5\% decrease in the PR of RGB-Event voxel, respectively. In contrast, our method makes the PR of RGB-Event voxel and RGB-Event frame drop by 70.7\% and 70.3\%, respectively, which almost completely disables the tracker's localization ability.

\begin{table}[t]
\centering
\caption{Performance Comparison of Attack Methods on Unimodal Inputs}
\label{sigle_modlity_attack_result}
\resizebox{\columnwidth}{!}{
\footnotesize 
\setlength{\tabcolsep}{3.5pt} 
\begin{tabular}{c|cccccccccc}
\hline
\multirow{2}{*}{Input Type} & \multirow{2}{*}{Method} & \multicolumn{3}{c}{COESOT} & \multicolumn{3}{c}{FE108} & \multicolumn{3}{c}{VisEvent} \\ 
\cline{3-5} \cline{6-8} \cline{9-11}
& & PR & NPR & SR & PR & NPR & SR & PR & NPR & SR \\ \hline
\multirow{6}{*}{Event Voxel} & Original & 12.6 & 15.8 & 15.4 & 0.74 & 0.23 & 1.45 & 4.2 & 2.9 & 3.1 \\
& Noise & 13.2 & 17.6 & 16.1 & 0.33 & 0.11 & 0.37 & 5.3 & 3.3 & 3.7 \\
& FGSM~\cite{goodfellow2014explaining} & 8.0 & 9.5 & 7.8 & 0.32 & 0.12 & 0.54 & 3.0 & 2.5 & 1.0 \\
& AE-ADV~\cite{lee2022adversarial} & 9.2 & 11.1 & 10.5 & 0.99 & 0.28 & 1.51 & 3.3 & 2.6 & 1.9 \\
& DARE-SNN~\cite{yao2024exploring} & 11.7 & 15.7 & 14.1 & 0.31 & 0.12 & 0.51 & 3.6 & 2.6 & 1.9 \\
& Ours & \textbf{6.9} & \textbf{8.2} & \textbf{7.2} & \textbf{0.30} & \textbf{0.10} & 0.40 & \textbf{2.8} & \textbf{2.4} & \textbf{0.9} \\ \hline
\multirow{4}{*}{Event Frame} & Original & 59.0 & 58.7 & 49.0 & 88.96 & 60.95 & 58.04 & 52.8 & 42.6 & 35.9 \\
& Noise & 47.0 & 47.3 & 39.2 & 71.64 & 49.42 & 45.15 & 44.1 & 34.1 & 26.5 \\
& FGSM~\cite{goodfellow2014explaining} & 54.6 & 55.1 & 45.9 & 79.68 & 53.31 & 51.26 & 49.7 & 40.5 & 33.9 \\
& Ours & \textbf{4.8} & \textbf{4.4} & \textbf{2.7} & \textbf{34.23} & \textbf{19.39} & \textbf{20.95} & \textbf{39.4} & \textbf{30.4} & \textbf{26.2} \\ \hline
\multirow{4}{*}{RGB Frame} & Original & 73.7 & 73.0 & 60.0 & 70.73 & 45.48 & 44.87 & 66.7 & 60.8 & 48.9 \\
& Noise & 21.9 & 18.3 & 12.3 & 1.99 & 0.95 & 0.76 & 30.5 & 19.4 & 12.1 \\
& FGSM~\cite{goodfellow2014explaining} & 7.0 & 6.7 & 4.3 & 0.39 & 0.12 & 0.35 & 7.0 & 4.6 & 2.8 \\
& Ours & \textbf{4.7} & \textbf{4.3} & \textbf{2.2} & \textbf{0.28} & \textbf{0.08} & \textbf{0.19} & \textbf{3.8} & \textbf{3.1} & \textbf{1.1} \\ \hline
\end{tabular}
}
\end{table}

\begin{table}[t]
\centering
\caption{Performance comparison of multimodal input attack methods}
\label{multimodal_attack_results}
\resizebox{\columnwidth}{!}{
\footnotesize
\setlength{\tabcolsep}{3.5pt}  
\begin{tabular}{c|cccccccccc}  
\hline
\multirow{2}{*}{\textbf{Input Type}} & 
\multirow{2}{*}{\textbf{Method}} & 
\multicolumn{3}{c}{\textbf{COESOT}} & 
\multicolumn{3}{c}{\textbf{FE108}} & 
\multicolumn{3}{c}{\textbf{VisEvent}} \\ 
\cline{3-5} \cline{6-8} \cline{9-11}  
 & & PR & NPR & SR & PR & NPR & SR & PR & NPR & SR \\ 
\hline
\multirow{5}{*}{RGB-Event Frames} 
 & Original & 74.5 & 73.7 & 61.4 & 90.33 & 64.64 & 60.04 & 65.3 & 58.7 & 47.6  \\
 & Noise & 37.9 & 40.0 & 31.8 & 44.57 & 29.32 & 27.05 & 37.6 & 24.6 & 18.8\\
 & FGSM~\cite{goodfellow2014explaining} & 40.5 & 43.0 & 37.3 & 76.44 & 29.32 & 45.97 & 41.2 & 28.4 & 25.8\\
 & Ours & \textbf{4.2} & \textbf{4.2} & \textbf{0.7} & \textbf{9.75} & \textbf{3.31} & \textbf{7.14} & \textbf{8.2} & \textbf{4.1} & \textbf{3.6}\\
\hline
\multirow{6}{*}{RGB-Event Voxel}
 & Original & 75.8 & 74.8 & 62.0 & 68.79 & 43.28 & 43.01 & 69.5 & 64.7 & 52.6  \\
 & Noise & 22.9 & 20.4 & 14.2 & 1.57 & 0.65 & 0.58 & 34.0 & 24.3 & 14.6  \\
 & FGSM~\cite{goodfellow2014explaining} & 6.9 & 6.4 & 4.7 & 0.50 & 0.13 & 0.28  & 7.1 & 5.6 & 4.1 \\ 
 & AE-ADV~\cite{lee2022adversarial} & 6.6 & 4.7 & 3.0 & 0.64 & 0.17 & 0.49 & 3.8 & 2.8 & 1.6 \\
 & DARE-SNN~\cite{yao2024exploring} & 6.3 & 4.6 & 2.8 & 0.57 & 0.16 & 0.41 & 4.1 & 3.0 & 1.8 \\
 & Ours & \textbf{5.1} & \textbf{4.4} & 4.2 & \textbf{0.23} & \textbf{0.08} & \textbf{0.12} & \textbf{3.5} & \textbf{2.7} & 2.1 \\
\hline
\end{tabular}%
}
\end{table}

\newcommand{\yes}{\ding{51}}  
\newcommand{\no}{\ding{55}}   

\begin{figure}[t]
    \centering
    \includegraphics[width=0.95\linewidth]{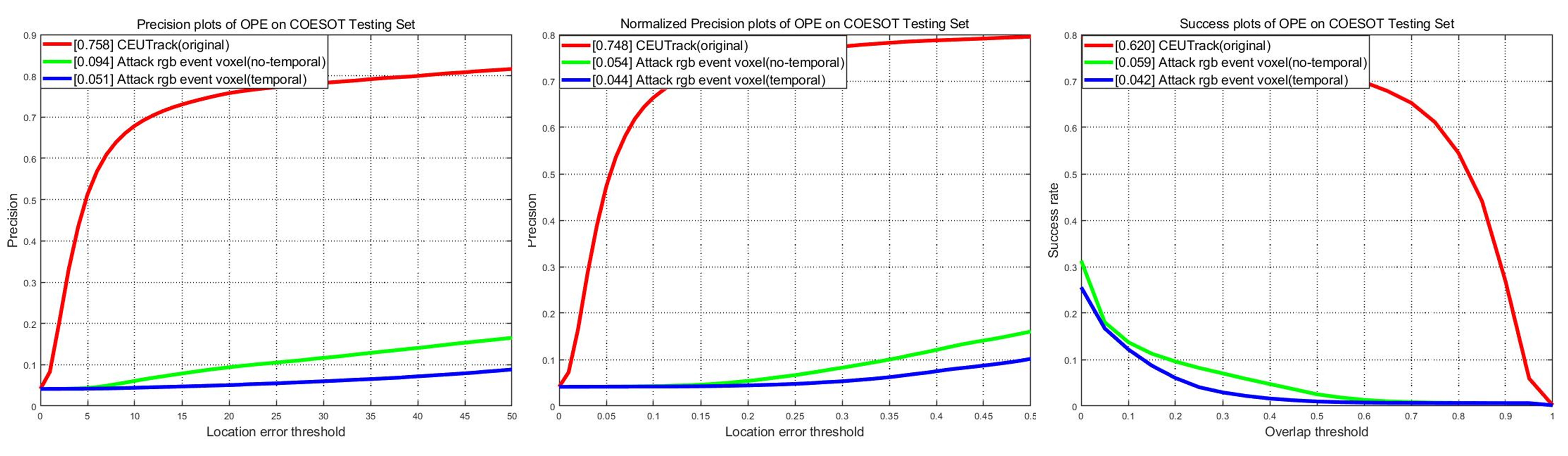} 
    \caption{Validation of Cross-Modal Attack Fusion for RGB-Event Voxel VOT with temporal perturbation on COESOT.}
    \label{fig:Validation of temporal perturbation}
\end{figure}

\begin{figure}[t]
    \centering
    \includegraphics[width=0.95\linewidth]{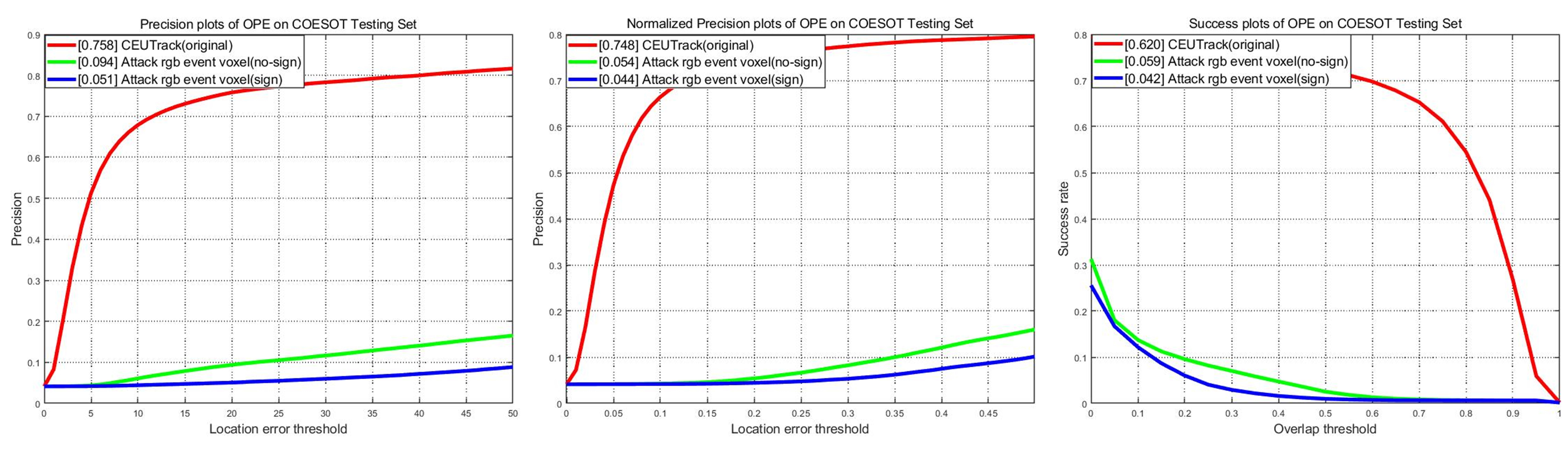} 
    \caption{Validation of Cross-Modal Attack Fusion for RGB-Event Voxel VOT's sign function on COESOT.}
    \label{fig:Validation of sign}
\end{figure}

\begin{figure}[t!]
    \centering
    \includegraphics[width=0.95\linewidth]{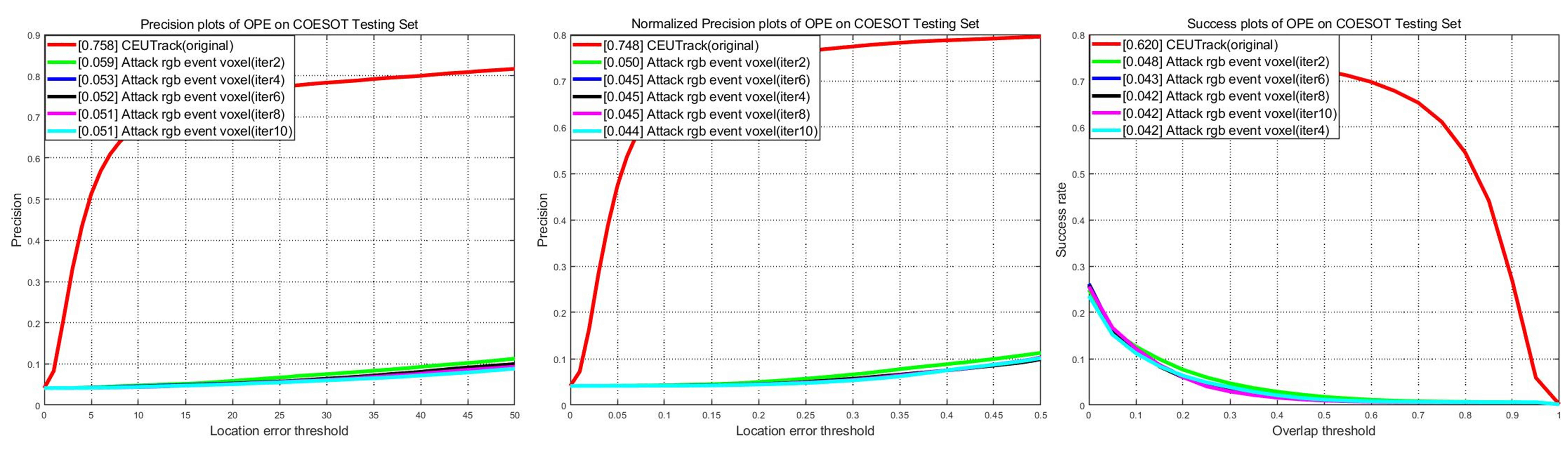} 
    \caption{Validation of Cross-Modal Attack Fusion for RGB-Event Voxel VOT with different iteration numbers on COESOT.}
    \label{fig:Validation of iteration numbers}
\end{figure}

\noindent $\bullet$ \textbf{Results on FE108 Dataset.~} In the unimodal attack scenario, FGSM~\cite{goodfellow2014explaining} causes the PR of Event voxels, Event frames and RGB frames to decrease by 0.42\%, 9.28\%, and 70.34\%, respectively. In contrast, AE-ADV~\cite{lee2022adversarial} and DARE-SNN~\cite{yao2024exploring} have less impact on the PR of Event voxels, which are reduced by 0.25\% and 0.43\%, respectively. However, our method performs more significantly in unimodal attacks, decreasing the PR of Event voxels, Event frames and RGB frames by 0.44\%, 54.73\% and 70.45\%, respectively. In the multimodal scenario, FGSM~\cite{goodfellow2014explaining} leads to a decrease in PR of 68.29\% and 13.89\% for RGB-Event voxels and RGB-Event frames, respectively.AE-ADV~\cite{lee2022adversarial} and DARE-SNN~\cite{yao2024exploring} reduce the PR of RGB-Event voxels by 68.15\% and 68.22\%, respectively. However, our method performs more prominently in multimodal attacks, leading to a decrease in PR of 68.56\% and 80.58\% for RGB-Event voxels and RGB-Event frames, respectively.

\noindent $\bullet$ \textbf{Results on VisEvent Dataset.~} In the unimodal attack scenario, the FGSM~\cite{goodfellow2014explaining} leads to a decrease in the PR of Event voxels, Event frames and RGB frames by 1.4\%, 3.1\%, and 59.7\%, respectively. In contrast, the AE-ADV~\cite{lee2022adversarial} and DARE-SNN~\cite{yao2024exploring} have less impact on the PR of Event voxels, which decrease by 0.9\% and 0.6\%, respectively. However, our method performs more significantly in unimodal attacks, causing the PR of Event voxels, Event frames, and RGB frames to decrease by 1.4\%, 13.4\%, and 62.9\%, respectively. In addition, our method has the largest reduction in NPR and SR.
In multimodal scenarios, the FGSM~\cite{goodfellow2014explaining} leads to a 62.4\% and 28.3\% decrease of PR for RGB-Event voxel and RGB-Event frame, respectively. The AE-ADV~\cite{lee2022adversarial} and DARE-SNN~\cite{yao2024exploring} attacks lead to a 69.2\% and 69.5\% decrease of PR for RGB-Event voxel, respectively. However, our method performs more prominently in the multimodal attack, causing the PR of the RGB-Event voxel and RGB-Event frame to drop by 65.5\% and 65.6\%, respectively, and almost completely disabling the tracker's localization ability.

\begin{figure}[t]
    \centering
    \includegraphics[width=\columnwidth]{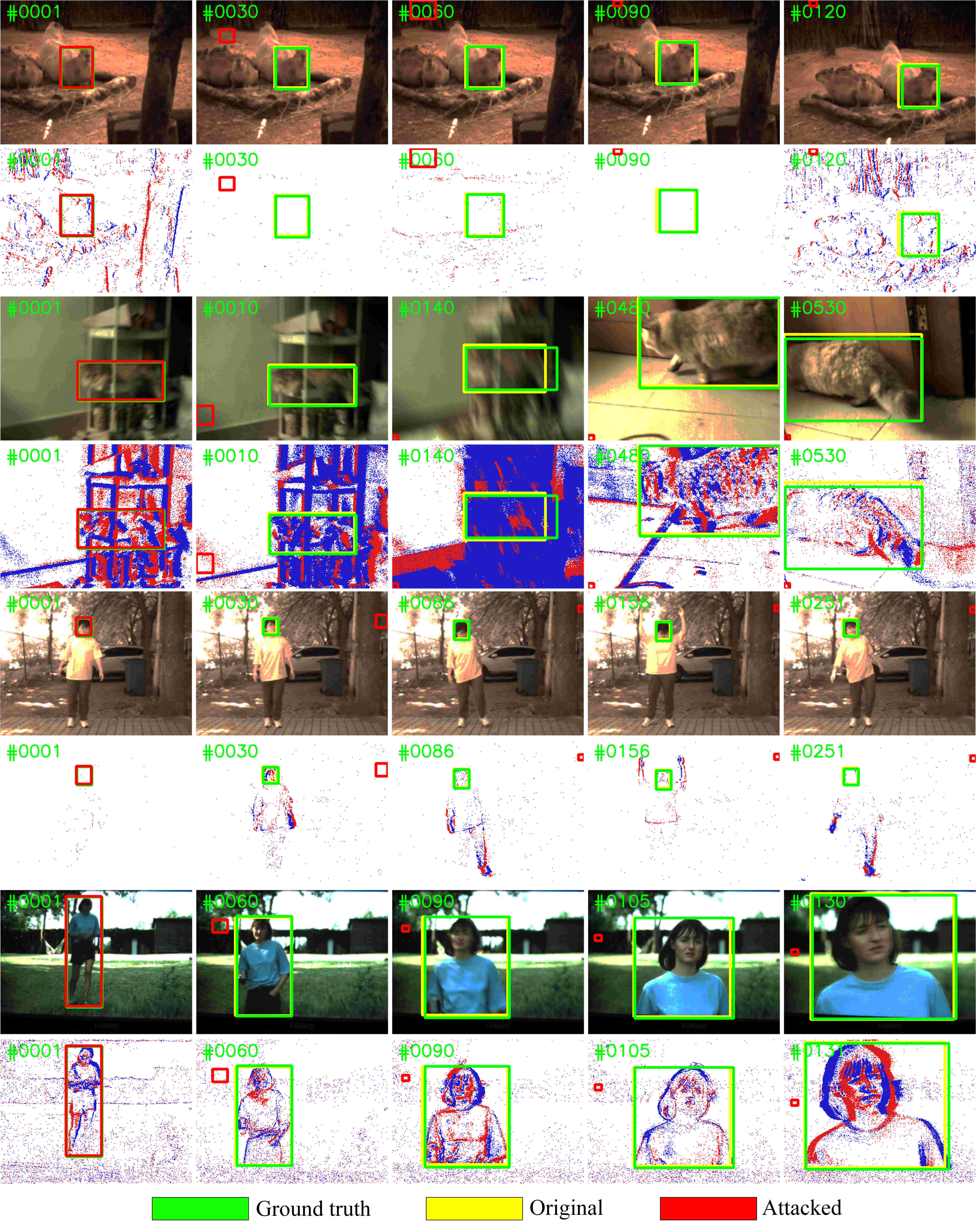} 
    \caption{Visualization results of the bounding box in the cross-modal attack of the RGB-Event voxels on the COESOT dataset. 
    Ground truth annotations (green boxes), original tracking results (yellow boxes), and attacked tracking results (red boxes) are overlaid on the RGB frames and Event frames.}  
    \label{fig:Visualization results of the bounding box}
\end{figure}

\subsection{Ablation Study}
In this section, we conduct ablation studies, including initial perturbation injection (AdvInit), gradient-guided optimization (GradOpt) strategy, and adversarial loss function. We conduct experiments on the COESOT dataset using the CEUTrack framework as a benchmark to verify the contribution of each module on the attack effectiveness.

\noindent $\bullet$ \textbf{Initial Perturbation Injection.~} Table~\ref{tab:Ablation Experiment of Configuration} demonstrates the impact of the AdvInit method on the performance of the CEUTrack tracker in the Event voxel scenario. By introducing the AdvInit method, the Event voxel-based CEUTrack tracker reduces the PR by 0.9\%. In Table~\ref{table: experiment of modality}, the AdvInit method reduces the PR by 0.2\% in the RGB Event voxel scenario. These experimental results demonstrate that our proposed AdvInit method successfully reduces the performance of the tracker through the strategy that injects voxels into the search patches.

\noindent $\bullet$ \textbf{Gradient-Guided Optimization.~} As shown in Tables ~\ref{tab:Ablation Experiment of Configuration} and~\ref{table: experiment of modality}, the GradOpt method makes the performance of the tracker degrade by optimizing the spatial location of Event voxel through gradient. For example, in the Event voxel unimodal scene, the GradOpt method reduces the PR by 5.4\%. In the RGB event voxel scene, the GradOpt method reduces the PR by 0.2\%. The above experimental results sufficiently demonstrate the effectiveness of the GradOpt approach in adversarial attacks in both unimodal and multimodal scenarios.

\begin{table}[t]
\centering
\caption{The Ablation Experiment of Configuration}
\label{tab:Ablation Experiment of Configuration}
\resizebox{\columnwidth}{!}{
\begin{tabular}{cccccccc}
\toprule
\multirow{2}{*}{\textbf{Input Type}} 
& \multicolumn{4}{c}{\textbf{Configuration}} 
& \multicolumn{3}{c}{\textbf{Score}} \\
\cmidrule(lr){2-5} \cmidrule(lr){6-8}
& \textbf{AdvInit} & \textbf{GradOpt} & \textbf{Track} & \textbf{Adv.} 
& \textbf{PR} & \textbf{NPR} & \textbf{SR} \\ 
\midrule
\multirow{6}{*}{Event Voxel}         
& \no  & \no  & \no  & \no  & 12.6 & 15.8 & 15.4 \\
& \yes & \no  & \no  & \no  & 11.7 & 15.7 & 14.1 \\
& \no  & \yes & \yes & \no  & 7.9  & 9.6  & 10.1 \\
& \no  & \yes & \no  & \yes & 7.2  & 8.7  & 8.2  \\
& \yes & \yes & \yes & \no  & 7.9  & 9.6  & 9.5  \\
& \yes & \yes & \no  & \yes & \textbf{6.9}  & \textbf{8.2}  & \textbf{7.2}  \\ 
\midrule 
\multirow{6}{*}{RGB Event Voxel}     
& \no  & \no  & \no  & \no  & 5.1 & 4.5 & 4.5 \\
& \yes & \no  & \no  & \no  & 5.1 & 4.5 & 4.5 \\
& \no  & \yes & \yes & \no  & 5.2 & 4.5 & 1.8 \\
& \no  & \yes & \no  & \yes & 5.1 & 4.5 & 4.2 \\
& \yes & \yes & \yes & \no  & 5.3 & 4.5 & 4.5 \\
& \yes & \yes & \no  & \yes & \textbf{5.1} & \textbf{4.4} & 4.2 \\ 
\midrule
\multirow{3}{*}{RGB Event Frame}     
& \no  & \no  & \no  & \no  & 74.5 & 73.7 & 61.4 \\
& \no  & \no  & \yes & \no  & 5.3  & 4.5  & 4.7  \\
& \no  & \no  & \no  & \yes & \textbf{4.2}  & \textbf{4.2}  & \textbf{0.7}  \\ 
\bottomrule
\end{tabular}
}
\end{table}

\begin{table}[t]
\centering
\caption{Ablation Study of Modality in Multimodal Attacks}
\label{table: experiment of modality}
\resizebox{\columnwidth}{!}{  
\footnotesize  
\setlength{\tabcolsep}{4.5pt}  
\begin{tabular}{@{}ccc*{4}{c}*{3}{c}@{}}  
\toprule
\multirow{2}{*}{\textbf{Input Type}} 
& \multicolumn{2}{c}{\textbf{Attacked Modality}} 
& \multicolumn{4}{c}{\textbf{Configuration}} 
& \multicolumn{3}{c}{\textbf{Score}} \\
\cmidrule(lr){2-3} \cmidrule(lr){4-7} \cmidrule(lr){8-10}
& \textbf{RGB} & \textbf{Event} 
& \textbf{AdvInit} & \textbf{GradOpt} & \textbf{Track} & \textbf{Adv.} 
& \textbf{PR} & \textbf{NPR} & \textbf{SR} \\ 
\midrule
\multirow{8}{*}{RGB-Event Voxel}
& \no & \no & \no & \no & \no & \no & 75.8 & 74.8 & 62.0 \\
& \yes & \no & \no & \no & \yes & \no & 5.2 & 4.5 & 1.8 \\
& \yes & \no & \no & \no & \no & \yes & \textbf{5.0} & \textbf{4.5} & 4.4 \\ 
\cmidrule{2-10}
& \no & \yes & \yes & \no & \no & \no & 75.6 & 74.6 & 61.7 \\
& \no & \yes & \no & \yes & \yes & \no & 75.1 & 74.1 & 61.2 \\
& \no & \yes & \no & \yes & \no & \yes & 75.0 & 74.0 & 61.2 \\
& \no & \yes & \yes & \yes & \yes & \no & 75.4 & 74.3 & 61.4 \\
& \no & \yes & \yes & \yes & \no & \yes & \textbf{75.0} & \textbf{74.0} & \textbf{61.2} \\ 
\midrule
\multirow{5}{*}{RGB-Event Frame}     
& \no & \no & \no & \no & \no & \no & 74.5 & 73.7 & 61.4 \\
& \yes & \no & \no & \no & \yes & \no & 5.3 & 4.6 & 4.6 \\
& \yes & \no & \no & \no & \no & \yes & \textbf{5.3} & \textbf{4.6} & \textbf{4.6} \\ 
\cmidrule{2-10}
& \no & \yes & \no & \no & \yes & \no & 22.0 & 19.9 & 16.8 \\
& \no & \yes & \no & \no & \no & \yes & \textbf{21.6} & \textbf{18.9} & \textbf{14.6} \\ 
\bottomrule
\end{tabular}
}
\end{table}

\noindent $\bullet$ \textbf{Adversarial Loss Function.~} Tables~\ref{tab:Ablation Experiment of Configuration} and~\ref{table: experiment of modality} show the performance comparison of our proposed adversarial loss function with the track loss function in different scenarios. In the unimodal scene, the track loss function decreases the PR with by 4.7\%, while our adversarial loss function decreases the PR with by 5.7\% under the same condition. In the multimodal scene, the track loss function decreases the PR with by 69.2\% and 70.3\%, respectively, while our adversarial loss function reduces the PR of 70.3\% and 70.7\%, respectively. The experimental results show that our proposed adversarial loss function further reduces the performance of the tracker in both unimodal and multimodal scenarios by introducing the results of the original model.

\noindent $\bullet$ \textbf{Temporal Perturbation.~} As shown in Fig.~\ref{fig:Validation of temporal perturbation}, we compare the attack effect of temporal perturbation in the RGB Event voxel fusion input scenario. The experimental results demonstrate that the tracking coherence of a visual object tracker over a time series can be significantly weakened by introducing temporal perturbations.

\noindent $\bullet$ \textbf{Analysis of Number of Iterations.~} 
In addition, we also conduct an in-depth analysis of the key parameters during the generation of adversarial samples. The results in Fig.~\ref{fig:Validation of iteration numbers} demonstrate that increasing the number of iterations can improve the attack performance of the adversarial samples, but it also consumes more computational resources and time. The attack performance gradually converges when the number of iterations reaches 10.

\noindent $\bullet$ \textbf{Analysis of Sign Function.~} 
In the process of generating the adversarial samples, the sign function can indicate the direction of perturbation generation by taking the sign function for the gradient. As shown in Fig.~\ref{fig:Validation of sign}, the performance of the attack using the sign function is significantly better than the no sign function, which indicates that the sign function plays an important role in improving the performance of the attack.

\begin{figure}[t]
\centering
\includegraphics[width=\columnwidth]{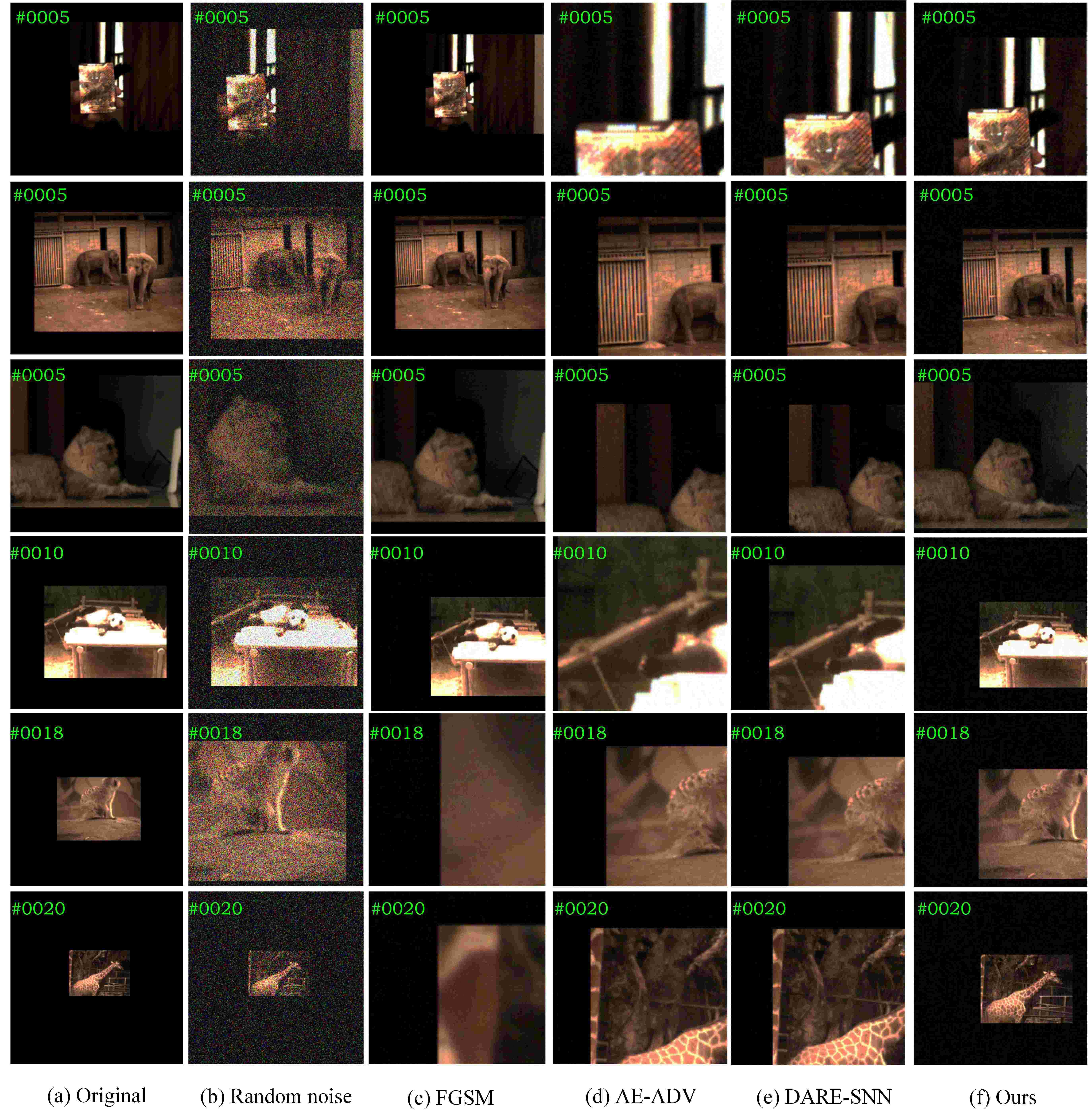} 
\caption{Demonstration of adversarial samples generated by different adversarial methods on RGB-Event voxel in the search patch on the COESOT Dataset, note that visualization cannot be performed due to the discreteness of event Voxel.}
\label{fig:Visualization of adversarial samples}
\end{figure}

\subsection{Visualization} 

\noindent $\bullet$ \textbf{Visualization of Tracking Results.~} Figure~\ref{fig:Visualization results of the bounding box} further demonstrates the effectiveness of our method's attack on different sequences. By comparing the Ground-truth annotations (green boxes), the original tracking results (yellow boxes), and the tracking results after the attack (red boxes), it can be seen that our method is able to quickly and significantly reduce the tracker's localization accuracy of the target object. The visual target tracker completely loses its ability to track the target location in subsequent frames over time. 

\noindent $\bullet$ \textbf{Adversarial Sample Presentations on RGB-Event voxel} \label{sec:vis} 
Figure~\ref{fig:Visualization of adversarial samples} shows the visualization results of the adversarial samples generated by the attack method against the RGB Event Voxel. Due to the non-visualization characteristic of Event Voxel, we only present the adversarial samples for the corresponding RGB frames. Random noise generates adversarial samples by sampling Gaussian noise, but lacks fine control for the noise. The FGSM~\cite{goodfellow2014explaining} method generates adversarial samples by one iteration, but its attack effect is relatively weak. 
AE-ADV and DARE-SNN attack the visual tracking model by modifying the timestamps and polarity~\cite{jang2016categorical} of the event data. However, these methods may lead to distortion of the target object. In contrast, our method utilizes AdvInit and GradOpt to generate perturbations, which achieves an effective attack on the visual tracking model while maintaining the relative integrity of the target object.

\begin{figure}[t]
\centering
\includegraphics[width=\columnwidth]{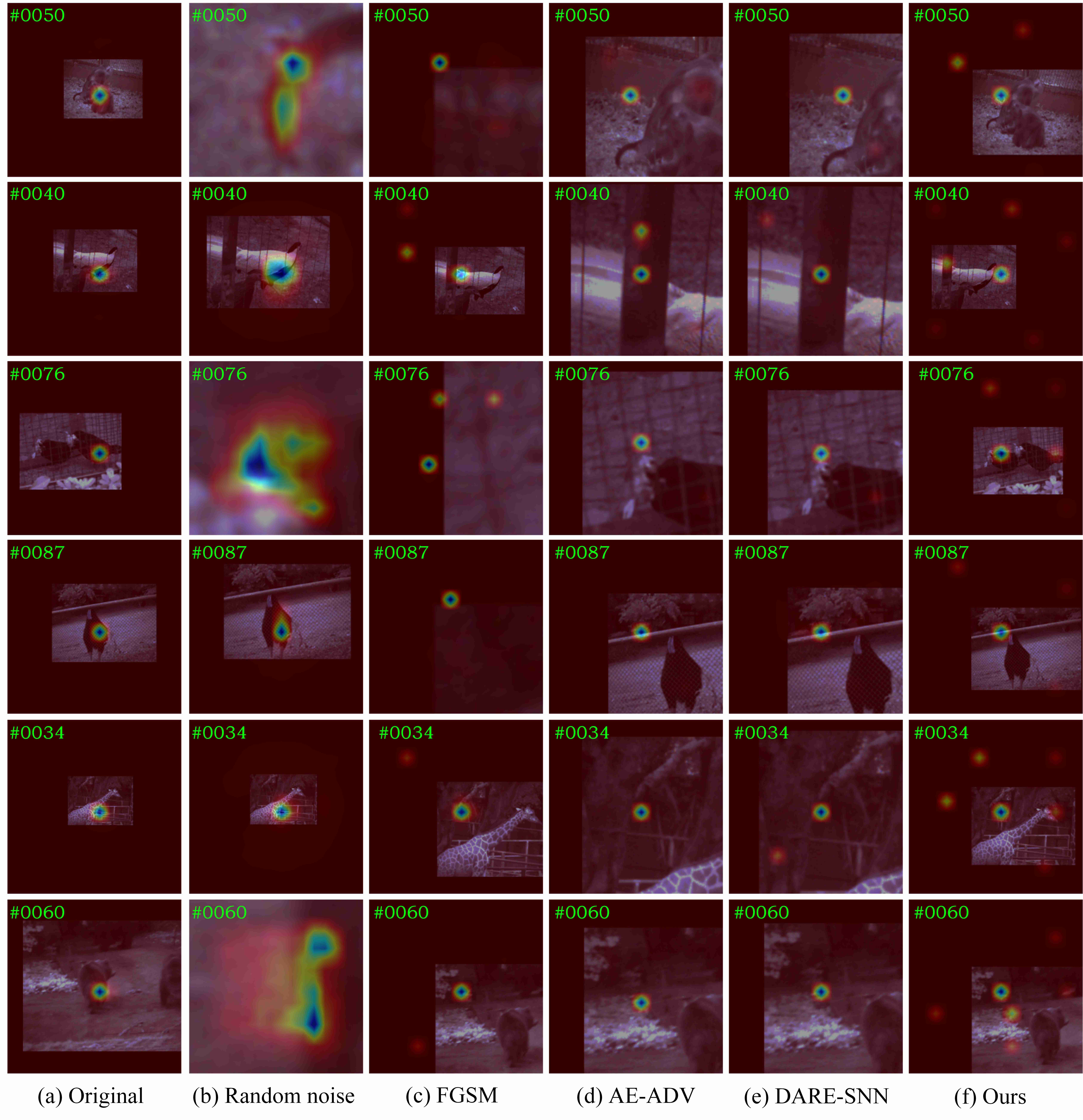} 
\caption{Comparison of activation maps with different adversarial attacks for the RGB-Event voxel tracking model on the COESOT dataset.}
\label{fig:Visualization of CAM}
\end{figure}

\begin{figure}
\centering
\includegraphics[width=\columnwidth]{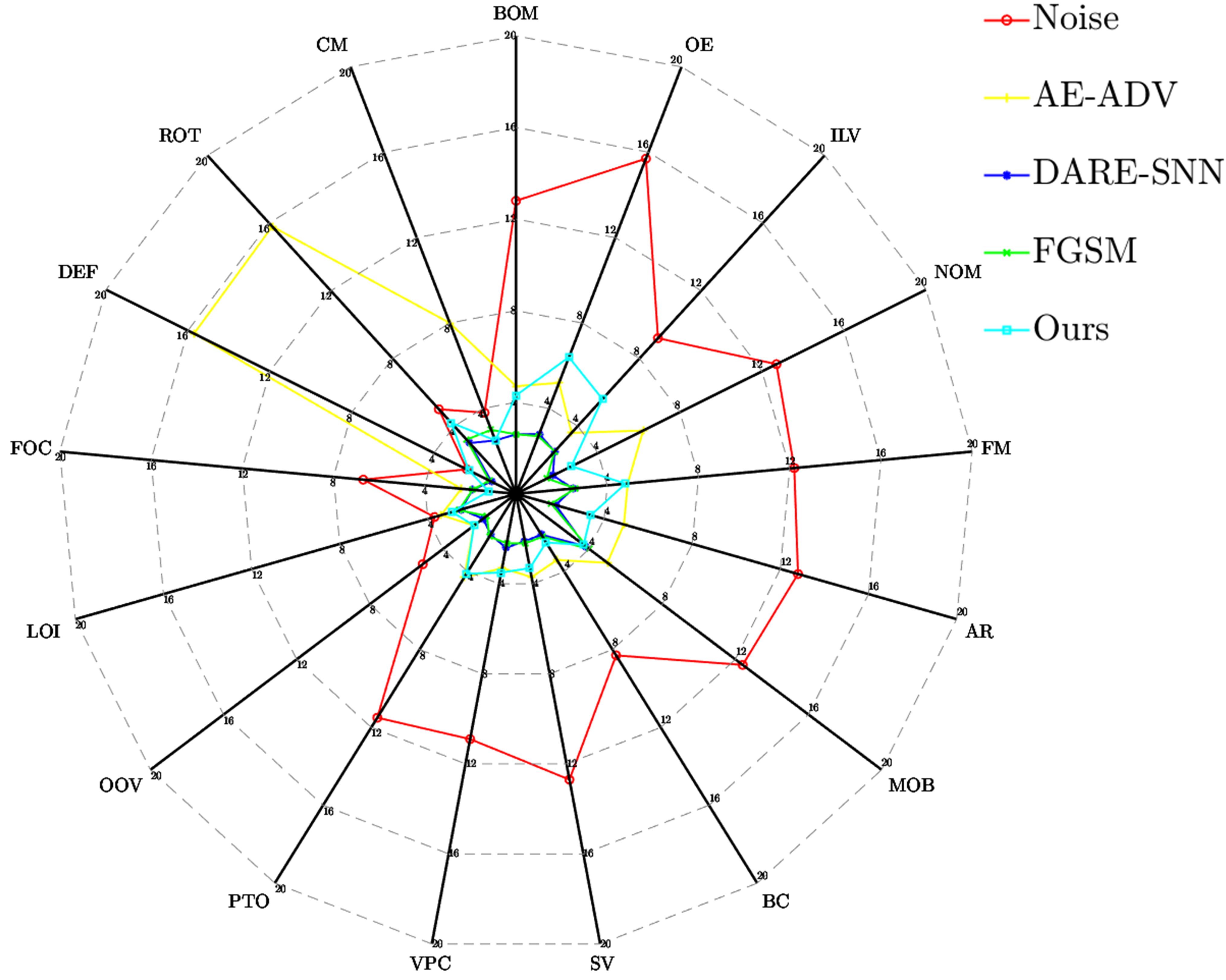} 
\caption{Success scores comparison of different attributes on COESOT dataset.}
\label{fig:Visualization of SR}
\end{figure}

\noindent $\bullet$ \textbf{Adversarial Sample Presentations on RGB-Event frame}. 
Figure~\ref{fig:Visualization of SR} illustrates the adversarial examples generated by our attack against an RGB event frame-based visual object tracker. By introducing multiple distractor targets, our method significantly decreases the tracker's accuracy. In contrast, perturbations generated by random noise not only yield limited attack performance, but also produce noticeable image artifacts, and FGSM~\cite{goodfellow2014explaining} fails to achieve a robust adversarial effect while preserving sample integrity. Furthermore, our approach achieves significantly improved adversarial performance while maintaining virtually imperceptible perturbations, demonstrating a distinct advantage over both random noise and the FGSM-based attack method.

\noindent $\bullet$ \textbf{Visualization of activation maps for adversarial examples on RGB-Event voxel}. 
Figure ~\ref{fig:Visualization of CAM} presents the activation maps generated by different adversarial attack methods on the COESOT dataset for the RGB-Event voxel-based visual object tracking model. Our method introduces the localization signals of multiple disturbing targets while keeping the adversarial examples undistorted, thus offsetting the position of the original target and perturbing the visual object tracking. The random noise method disrupts the visual representation of the original image by adding noise to the image, resulting in significant changes in the activation maps for the visual object tracker with target disturbances. FGSM~\cite{goodfellow2014explaining}, AE-ADV~\cite{lee2022adversarial}, and DARE-SNN~\cite{yao2024exploring} by perturbing the location of the target and causing the tracker to lose track. In contrast, our method achieves a better attack effect by introducing more perturbed targets.

\begin{figure}[t]
\centering
\includegraphics[width=\columnwidth]{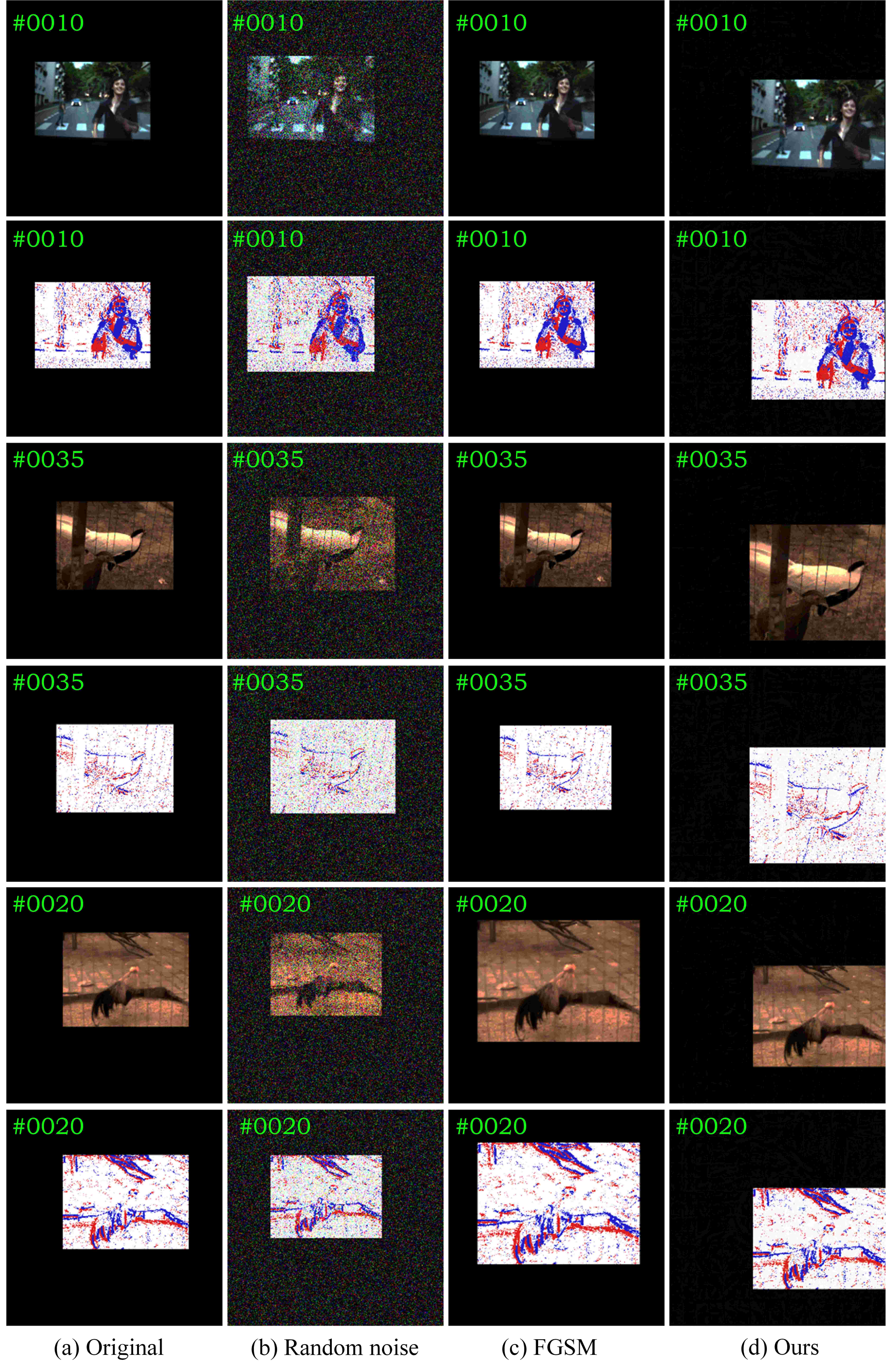} 
\caption{Visualization of different adversarial examples against RGB-Event frame attacks on the COESOT dataset.}
\label{fig:Visualization of FrameAdv}
\end{figure}

\section{Conclusion and Future Works} 
In this paper, we explore the robustness of RGB-Event-based visual object tracking models under adversarial attacks. We reveal the sensitivity of RGB-Event under cross-modal attack for the perturbations and propose a cross-modal attack algorithm for RGB-Event visual object tracking. Firstly, the algorithm generates RGB frame adversarial samples by optimizing the perturbation through the adversarial loss, then the final adversarial samples of Event voxels are generated by injecting Event voxels into the target region as initialized Event voxel adversarial samples, and the adversarial samples of the ultimate Event voxels are generated by optimizing the spatial locations of the Event voxels through the gradient information. In addition, we also explore the robustness of the RGB-Event Frame under adversarial attacks by integrating the gradient information of the RGB-Event frame to optimize the cross-modal universal perturbation to generate adversarial samples. In the future, we plan to explore more RGB-Event-based visual tracking model architectures to further reveal the security of RGB-Event data under adversarial attacks. We expect that this research will motivate attention to the adversarial robustness of multi-modal model trackers and consider the security of trackers when designing multi-modal trackers.



\small{ 
\bibliographystyle{IEEEtran}
\bibliography{reference}
}


\clearpage
\appendix
In the Supplementary Material, we introduce unimodal-based attack algorithms in sec.~\ref{sec:unimodal}, AE-ADV~\cite{lee2022adversarial} and DARE-SNN~\cite{yao2024exploring} attack algorithms in sec.~\ref{sec:AE-ADV} and sec.~\ref{sec:DARE-SNN}.
\section{Unimodal-based attack algorithms} 
\label{sec:unimodal}

\subsection{Algorithms for adversarial attacks on RGB Frame-based visual object tracking}

We use the PGD algorithm-based approach to attack the RGB Frame-based visual object tracking model to generate adversarial samples. Different from the traditional PGD algorithm, we introduce the adversarial loss function proposed in the main paper when generating the adversarial samples, and combine the temporal perturbation strategy to disrupt the visual object tracker. The specific algorithm process is shown in algorithm~\ref{alg:rgb_attack}. In addition, an adversarial attack for the Event Frame-based visual object tracking model is conducted in similar way to the RGB Frame-based model, the only difference is the different modality of the input data.
\begin{algorithm}[h]
\caption{Algorithm for AE-ADV Attack}
\label{alg:AEADV}
\raggedright  
\textbf{Input}: Original RGB Frames ${I}$, Original Event Voxel $V=\{(x_k,y_k,t_k,p_k)\}_{k=1}^{N}$, Target location $S(x,y,w,h)$, Perturbation $\eta_{\text{rgb}}$, $\eta_{\text{event}}$, true label $gt$ \\
\textbf{Parameter}: Number of total iteration $M$, step size $\alpha$, perturbation size $\epsilon$,model parameters $\theta$  \\
\textbf{Output}: Adversarial sample ${I}^{\ast}$ and $V^{\ast}$
\vspace{-5mm}
\begin{algorithmic}[1]
    \FOR{$t = 1$ to $T$}
        \STATE Retrieve the RGB frame ${I}^{t}$ and Event voxel $V^{t}$ within the search space at the current temporal instance
        \STATE Get invalid number of voxel $N_v$ 
        \STATE ${V^{t}}' \leftarrow $ Add $N_v$ voxel in the target region $S(x,y,w,h)$
        \STATE Update ${I}^{t},{V^{t}}'$ according to $\eta_{\text{rgb}},\eta_{\text{event}}$ 
        \FOR{$i = 1$ to $M$}
            \STATE Compute $R_{rgb}=\nabla J({I_{M}^{t}},\theta)$ 
            \STATE ${I}^{\ast} \leftarrow \text{Proj}_{\epsilon}\left({I_{M}^{t}} + \alpha \cdot \text{sign}(R_{rgb})\right)$  
            \STATE Update $\eta_{\text{rgb}} \leftarrow {I^{\ast }} - I$
        
            \STATE Compute temporal gradient $R_{event} = \nabla J({V^{t}_{M}}', \theta)$ 
            \STATE $V^{\ast} \leftarrow \text{Proj}_{\epsilon}\left({V^{t}_{M}}' + \alpha \cdot \text{sign}(R_{event})\right)$
        \ENDFOR
        \STATE Update $\eta_{\text{event}} \leftarrow V^{\ast} - V$
    \ENDFOR
    \STATE Return $I^{\ast}$, $V^{\ast}$
\end{algorithmic}
\end{algorithm}
\subsection{AE-ADV Attack Algorithm}
\label{sec:AE-ADV}
This section details the AE-ADV~\cite{lee2022adversarial} adversarial attack method, which targets the raw Event stream data initialized by injecting null events in the event stream, then using the gradient information to modify the event timestamps to generate adversarial Event data with temporal perturbations. We migrate the AE-ADV attack framework to the RGB-Event voxel visual object tracking scenario as our comparison method. Under the multimodal attack paradigm, we adopt the PGD algorithm uniformly to implement the adversarial sample generation for RGB image modalities in order to ensure the consistency of the multimodal attack strategy. The specific implementation process is shown in algorithm~\ref{alg:AEADV}.
\begin{algorithm}[h]
\caption{Algorithm for DARE-SNN attack}
\label{alg:DARESNN}
\raggedright  
\textbf{Input}: Original RGB Frames ${I}$, Original Event Voxel $V=\{(x_k,y_k,t_k,p_k)\}_{k=1}^{N}$, Target location $S(x,y,w,h)$, Perturbation $\eta_{\text{rgb}}$, $\eta_{\text{event}}$, true label $gt$ \\
\textbf{Parameter}: Number of total iteration $M$, step size $\alpha$, perturbation size $\epsilon$,model parameters $\theta$  \\
\textbf{Output}: Adversarial sample ${I}^{\ast}$ and $V^{\ast}$
\vspace{-5mm}
\begin{algorithmic}[1]
    \FOR{$t = 1$ to $T$}
        \STATE Retrieve the RGB frame ${I}^{t}$ and Event voxel $V^{t}$ within the search patch at the current temporal instance
        \STATE Get invalid number of voxel $N_v$ 
        \STATE ${V^{t}}' \leftarrow $ Add $N_v$ voxel in the target region $S(x,y,w,h)$
        \STATE  Initialize the polarity p of Event voxel using Gumbel-Softmax~\cite{jang2016categorical}
        \STATE Update ${I}^{t},{V^{t}}'$ according to $\eta_{\text{rgb}},\eta_{\text{event}}$ 
        \FOR{$i = 1$ to $M$}
            \STATE Compute $R_{rgb}=\nabla J({I_{M}^{t}},\theta)$ 
            \STATE ${I}^{\ast} \leftarrow \text{Proj}_{\epsilon}\left({I_{M}^{t}} + \alpha \cdot \text{sign}(R_{rgb})\right)$  
            \STATE Update $\eta_{\text{rgb}} \leftarrow {I^{\ast }} - I$
            \STATE Compute gradients of polarity $p$ $R_{p} = \nabla J({V^{t}_{M}}', \theta)$ 
            \STATE Updating Event voxel polarity with $R_{p}$
            \STATE Update $\eta_{\text{event}} \leftarrow V^{\ast} - V$
        \ENDFOR
    \ENDFOR
    \STATE Return $I^{\ast}$, $V^{\ast}$
\end{algorithmic}
\end{algorithm}

\subsection{DARE-SNN attack algorithm}
\label{sec:DARE-SNN}
In this section, we introduce the DARE-SNN~\cite{yao2024exploring} attack method, which is an adversarial attack algorithm for the original Event Stream. The method is initialized by injecting the attributes of the attacking category into the current category and subsequently optimizing the event stream polarity using gradient information. We migrate the core algorithm of the DARE-SNN attack to RGB-Event voxel visual object tracking and use it as our comparison method. In order to maintain the consistency in the multimodal attack, we adopt the PGD algorithm to generate the adversarial samples of RGB frame. The specific algorithm process is shown in algorithm~\ref{alg:DARESNN}.
\begin{algorithm}[h]
\caption{Algorithms for adversarial attacks on RGB Frame-based visual object tracking}
\label{alg:rgb_attack}
\raggedright
\textbf{Input}: Original frames $I$, pre-trained tracker $T$, target location $S(x,y,w,h)$, perturbation $\eta$, true label $gt$ \\
\textbf{Parameter}: number of total iteration $M$, step size $\alpha$, perturbation size $\epsilon$, and the model's parameters $\theta$ \\
\textbf{Output}: Adversarial frames ${I^{*}}$
\vspace{-5mm}
\begin{algorithmic}[1]
    \FOR{$t = 1$ to $T$}
        \STATE Get the frame $I^{t}$ in current search region
        \STATE Initialize adversarial frame ${I^{t}}' \leftarrow I^{t} + \eta$
        \FOR{$i = 1$ to $M$}
            \STATE Compute $R = \nabla J({I^{t}_M}', \theta)$ according to $\eta_{\text{rgb}},\eta_{\text{event}}$ 
            \STATE ${I^{*}} \leftarrow \text{Proj}_{\epsilon}\left({I^{t}_M}' + \alpha \cdot \text{sign}(R)\right)$
        \ENDFOR
        \STATE Update $\eta \leftarrow {I^{*}} - I$
    \ENDFOR
    \STATE Return ${I^{*}}$ and $\eta$
\end{algorithmic}
\end{algorithm}

\end{document}